\documentclass[11pt,a4paper]{article}
\usepackage[hyperref]{acl2017}
\usepackage{times}
\usepackage{latexsym}

\usepackage{url}
\usepackage{graphicx}
\usepackage{color}
\usepackage{amsmath}
\usepackage{bm}
\usepackage{booktabs}
\usepackage{multirow}
\usepackage{bm}
\usepackage{dblfloatfix}    
\usepackage{capt-of}
\usepackage{amssymb}
\usepackage{mathabx}
\usepackage [autostyle, english = american]{csquotes}
\usepackage{tabularx}
\usepackage{subcaption}

\usepackage{xcolor}
\usepackage{tikz}
\usepackage{pgfplots, pgfplotstable}
\usetikzlibrary{patterns}

\usepackage{verbatim}
\MakeOuterQuote{"}

\newcommand{\vt}[1]{\bm{\mathbf{#1}}}

\aclfinalcopy 
\def\aclpaperid{3246} 

\title{Learning Topic-Sensitive Word Representations}

\author{Marzieh Fadaee \qquad Arianna Bisazza \qquad Christof Monz\\
Informatics Institute, University of Amsterdam\\
Science Park 904, 1098 XH Amsterdam, The Netherlands\\
  {\tt \{m.fadaee,a.bisazza,c.monz\}@uva.nl} \\
 }

\date{}

\begin{document}
\maketitle
\begin{abstract}
Distributed word representations are widely used for modeling words in NLP tasks. Most of the existing models generate one representation per word and do not consider different meanings of a word.   
We present two approaches to learn multiple topic-sensitive representations per word by using Hierarchical Dirichlet Process. We observe that by modeling topics and integrating topic distributions for each document  we obtain representations that are able to distinguish between different meanings of a given word.
Our models yield statistically significant improvements for the lexical substitution task 
indicating that commonly used single word representations, even when combined with contextual information, are insufficient for this task. 
\end{abstract}

\section{Introduction}

Word representations in the form of dense vectors, or word embeddings, capture semantic and syntactic information \cite{mikolov2013efficient,pennington2014glove} and are widely used in many NLP tasks \cite{zou2013bilingual,levy2014dependency,tang2014learning,gharbieh2016word}.   
Most of the existing models generate one representation per word and do not distinguish between different meanings of a word. However, many tasks can benefit from using multiple representations per word to capture polysemy \cite{reisinger-mooney:2010:NAACLHLT}.
There have been several attempts to build repositories for word senses \cite{miller1995wordnet,navigli2010babelnet}, but this is laborious and limited to few languages. Moreover, defining a universal set of word senses is challenging as polysemous words can exist at many levels of granularity \cite{journals/lre/Kilgarriff97,Navigli2012}. 

In this paper, we introduce a model that uses a nonparametric Bayesian model, Hierarchical Dirichlet Process (HDP), to learn multiple topic-sensitive representations per word.  
\newcite{yao2011nonparametric} show that HDP is effective in learning topics yielding state-of-the-art performance for sense induction.  
 However, they assume that topics and senses are interchangeable, and train individual models per word making it difficult to scale to large data. 
Our approach enables us to use HDP to model senses effectively using large unannotated training data. 
We investigate to what extent distributions over word senses can be approximated by distributions over topics without assuming these concepts to be identical.
The contributions of this paper are: (i) We propose three unsupervised, language-independent approaches to approximate senses with topics and learn multiple topic-sensitive embeddings per word.
(ii) We show that in the Lexical Substitution ranking task  
our models outperform two competitive baselines.  

  \begin{figure*}[htb!]      
  \centering 
  \includegraphics[scale=0.43]{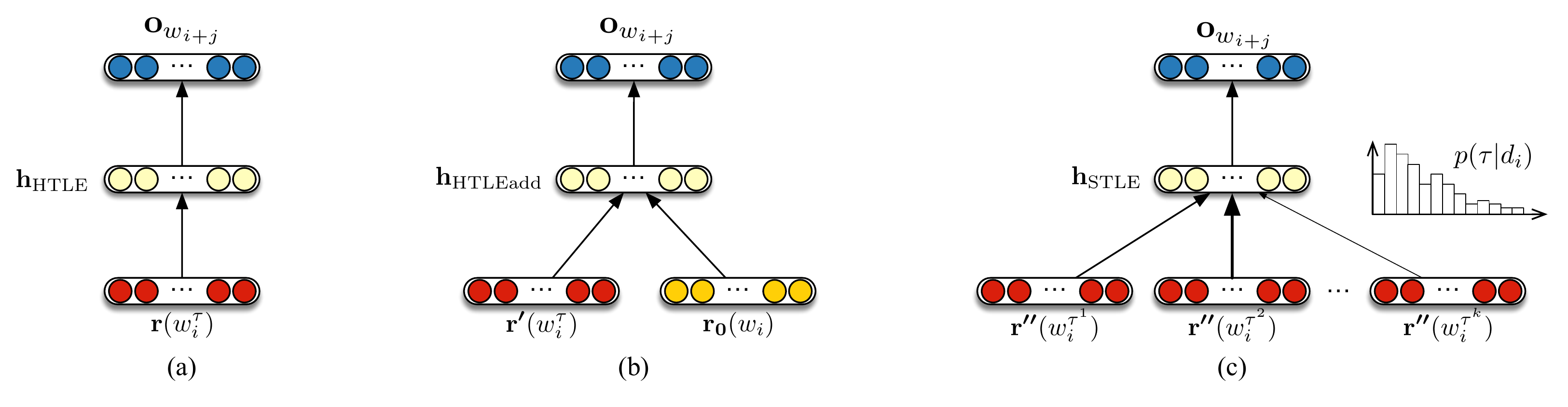} 
  \caption{Illustration of our topic-sensitive representation models: (a) hard-topic labeled representations 
  (HTLE),  
  (b) hard topic-labeled representations plus generic word representation 
  (HTLEadd),  
  (c) soft topic-labeled representations 
  (STLE). 
  }
    \label{modelsfig}
\end{figure*}

\section{Topic-Sensitive Representations}
\label{sect:model}

In this section we describe the proposed models. To learn topics from a corpus we use HDP \cite{teh2006hierarchical,lau2014learning}.  
The main advantage of this model compared to non-hierarchical methods like the Chinese Restaurant Process (CRP) is that 
each document in the corpus is modeled using a mixture model with  
topics shared between all documents \cite{citeulike:635668,brody-lapata:2009:EACL}. 
HDP yields two sets of distributions that we use in our methods: distributions over topics for words in the vocabulary, and distributions over topics for documents in the corpus. 

Similarly to \newcite{neelakantan2014efficient}, we use neighboring words  
to detect the meaning of the context, however,
we also use the two HDP distributions. By doing so, we take advantage of the topic of the document beyond the scope of the neighboring words, which is helpful when the immediate context of the target word is not sufficiently informative. We modify the Skipgram model \cite{mikolov2013efficient} to obtain multiple topic-sensitive representations per word type using topic distributions.
 In addition, we do not cluster context windows and train for different senses of the words individually. This  
reduces the sparsity problem and provides a better representation estimation for rare words.
We assume that meanings of words can be determined by their contextual information and use the distribution over topics to differentiate between occurrences of a word in different contexts, i.e., documents with different topics.
We propose three different approaches (see Figure~\ref{modelsfig}):
two methods with hard topic labeling of words and one with soft labeling.
  
\subsection{Hard Topic-Labeled Representations}
\label{sect:hardReps}

The trained HDP model can be used to hard-label a new corpus with one topic per word through sampling.
Our first model variant (Figure~\ref{modelsfig}(a)) relies on hard labeling by simply considering each word-topic pair as a separate vocabulary entry.
To reduce sparsity on the context side and share the word-level information between similar contexts, we use topic-sensitive representations for target words (input to the network) and standard, i.e., unlabeled, word representations for context words (output). Note that this results in different input and output vocabularies.
The training objective is then to maximize the log-likelihood of context words $w_{i+j}$ given the target word-topic pair $w_{i}^{\tau}$:
\begin{align*}\small
\mathcal{L}_{HardT\textrm{-}SG} = \frac{1}{I} \sum_{i=1}^I \sum_{\substack{-c \le j \le c \\ j \ne 0}}{\log p(w_{i+j} | w^{\tau}_{i})}
\end{align*}
\noindent%
where $I$ is the number of  
words in the training corpus, $c$ is the context  
size and $\tau$ is the topic assigned to $w_i$ by HDP sampling.
The embedding of a word in context $\vt{h}({w_i})$ is obtained by simply extracting the row of the input lookup table (\vt{r}) corresponding to the HDP-labeled word-topic pair:
\begin{equation}\label{eq:HTLE}
\vt{h}_\textrm{HTLE}({w_i}) = \vt{r}(w_i^{\tau})
\end{equation}

A possible shortcoming of the HTLE model is that the representations are trained separately and information is not shared between different topic-sensitive representations of the same word. 
To address this issue, we introduce a model variant that learns multiple topic-sensitive word representations and generic word representations simultaneously (Figure~\ref{modelsfig}(b)). 
In this variant (HTLEadd), the target word embedding is obtained by adding the word-topic pair representation ($\vt{r'}$) to the generic representation of the corresponding word ($\vt{r_0}$):
\begin{equation}\label{eq:HTLEadd}
\vt{h}_\textrm{HTLEadd}({w_i}) = \vt{r'}(w_i^\tau) + \vt{r_0}(w_i)
\end{equation}

\subsection{Soft Topic-Labeled Representations}
\label{sect:softReps}

 \begin{table*}[htb!]
\centering\small
\begin{tabular}{lll}
\toprule
word                 & \multicolumn{1}{l}{}        & \textbf{bat}                                                                                         \\ \midrule
Pre-trained w2v & \multicolumn{1}{l|}{}        & bats, batting, Pinch\_hitter\_Brayan\_Pena, batsman, batted, Hawaiian\_hoary, Batting       \\
Pre-trained GloVe    & \multicolumn{1}{l|}{}        & bats, batting, Bat, catcher,  fielder, hitter, outfield, hitting, batted, catchers, balls   \\
SGE  on Wikipedia              & \multicolumn{1}{l|}{}        & uroderma, magnirostrum, sorenseni, miniopterus, promops, luctus, micronycteris              \\
\hline
\multirow{2}{*}{TSE  on Wikipedia}            & \multicolumn{1}{l|}{ $\tau_1$} & ball, pitchout, batter, toss-for, umpire, batting, bowes, straightened, fielder, flies      \\
                     & \multicolumn{1}{l|}{$\tau_2$} & vespertilionidae, heran, hipposideros, sorenseni, luctus, coxi, kerivoula, natterer         \\ 
                     \midrule
word                 &                \multicolumn{1}{l}{}                  & \textbf{jaguar}                                                                                      \\ \midrule
Pre-trained w2v & \multicolumn{1}{l|}{}        & jaguars, Macho\_B, panther, lynx, rhino, lizard, tapir, tiger, leopard, Florida\_panther    \\
Pre-trained GloVe    & \multicolumn{1}{l|}{}        & jaguars, panther, mercedes, Jaguar, porsche, volvo, ford, audi, mustang, bmw, biuck         \\
SGE on Wikipedia           & \multicolumn{1}{l|}{}        & electramotive, viper, roadster, saleen, siata, chevrolet, camaro, dodge, nissan, escort     \\
\hline
\multirow{2}{*}{TSE  on Wikipedia}                   & \multicolumn{1}{l|}{$\tau_1$} & ford, bmw, chevrolet, honda, porsche, monza, nissan, dodge, marauder, peugeot, opel         \\
                     & \multicolumn{1}{l|}{$\tau_2$} & wiedii, puma, margay, tapirus, jaguarundi, yagouaroundi, vison, concolor, tajacu            \\ 
                     \midrule
word                 & \multicolumn{1}{l}{}        & \textbf{appeal}                                                                                      \\ \midrule
Pre-trained w2v & \multicolumn{1}{l|}{}        &       appeals, appealing, appealed, Appeal, rehearing, apeal, Appealing, ruling, Appeals \\
Pre-trained GloVe    & \multicolumn{1}{l|}{}        &      appeals, appealed, appealing, Appeal, court, decision, conviction, plea, sought    \\
SGE  on Wikipedia                    & \multicolumn{1}{l|}{}        &     court, appeals, appealed, carmody, upheld, verdict, jaruvan, affirmed, appealable   \\
\hline
\multirow{2}{*}{TSE  on Wikipedia}                   & \multicolumn{1}{l|}{$\tau_1$} &   court, case, appeals, appealed, decision, proceedings, disapproves, ruling\\
                     & \multicolumn{1}{l|}{$\tau_2$} &    sfa, steadfast, lackadaisical, assertions, lack, symbolize, fans, attempt, unenthusiastic    \\ \bottomrule 
\end{tabular}
\caption{Nearest neighbors of three examples in different representation spaces using cosine similarity. \textbf{w2v} and \textbf{GloVe} are pre-trained embeddings from \protect\cite{mikolov2013efficient} and \protect\cite{pennington2014glove} respectively. \textbf{SGE} is the Skipgram baseline and \textbf{TSE} is our topic-sensitive Skipgram (cf. Equation~(\ref{eq:HTLE})), both trained on Wikipedia. $\tau_k$ stands for HDP-inferred topic $k$.}
\label{embs}
\end{table*}

The model variants above rely on the hard labeling resulting from HDP sampling.
As a soft alternative to this, we can directly  
include the topic distributions estimated by HDP for each document (Figure~\ref{modelsfig}(c)). 
Specifically, for each update, we use the topic distribution to compute a weighted sum over the word-topic representations ($\vt{r''}$):
\begin{equation}\label{eq:STLE}
\vt{h}_\textrm{STLE}(w_i) = \sum_{k=1}^{T}  \ p(\tau_k | d_i) \ \vt{r''}(w_i^{\tau_k})
\end{equation}
where $T$ is the total number of topics, $d_i$ the document containing $w_i$, and $p(\tau_k | d_i)$ the probability assigned to topic $\tau_k$ by HDP in document $d_i$.
The training objective for this model is:
\begin{align*}\small
\mathcal{L}_{SoftT\textrm{-}SG} = \frac{1}{I} \sum_{i=1}^I \sum_{\substack{-c \le j \le c \\ j \ne 0}}{\log p(w_{i+j} | w_i, \tau)}
\end{align*}

\noindent%
where $\tau$ is the topic of document $d_i$ learned by HDP. 
The STLE model has the advantage of directly applying the distribution over topics in the Skipgram model. In addition, for each instance, we update all topic representations of a given word with non-zero probabilities, which has the potential to reduce the sparsity problem.

\subsection{Embeddings for Polysemous Words}

The representations obtained from our models are expected to capture the meaning of a word in different topics.
We now investigate whether these representations can distinguish between different word senses.
Table~\ref{embs} provides examples of nearest neighbors. For comparison we include
our own baseline, i.e., embeddings learned with Skipgram on our corpus, as well as Word2Vec \cite{mikolov2013distributed} 
and GloVe embeddings \cite{pennington2014glove} 
pre-trained on large data.

In the first example, the word \textit{bat} has two different meanings: animal or sports device. One can see that the nearest neighbors of the baseline and pre-trained word representations either center around one primary, i.e., most frequent, meaning of the word, or it is a mixture of different meanings. 
The topic-sensitive representations, on the other hand, correctly distinguish between the two different meanings. A similar pattern is observed for the word \textit{jaguar} and its two meanings: car or animal.
The last example, \textit{appeal}, illustrates a case where topic-sensitive embeddings are not clearly detecting different meanings of the word, despite having some correct words in the lists.  
Here, the meaning \textit{attract} does not seem to be captured by any embedding set. 

These observations suggest that topic-sensitive representations capture different word senses to some extent.
To provide a systematic validation of our approach, we now investigate whether topic-sensitive representations can improve tasks where polysemy is a known issue. 

\section{Evaluation}

In this section we present the setup for our experiments and empirically evaluate our approach on the context-aware word similarity and lexical substitution tasks.

\subsection{Experimental setup}

All word representations are learned on the English Wikipedia corpus containing 4.8M documents (1B tokens).
The topics are learned on a 100K-document subset of this corpus using the HDP implementation of \newcite{teh2006hierarchical}.
Once the topics have been learned, we run HDP on the whole corpus to obtain the word-topic labeling (see Section~\ref{sect:hardReps}) and the document-level topic distributions (Section~\ref{sect:softReps}).
We train each model variant with window size $c=10$ and different embedding sizes (100, 300, 600) initialized randomly. 

We compare our models to several baselines: Skipgram (SGE) and the best-performing multi-sense embeddings model per word type (MSSG) \cite{neelakantan2014efficient}. All model variants are trained on the same training data with the same settings, following suggestions by \newcite{mikolov2013efficient} and \newcite{levy2015improving}. For MSSG we use the best performing similarity measure (avgSimC) as proposed by \newcite{neelakantan2014efficient}.

\subsection{Context-Aware Word Similarity Task}

Despite its shortcomings \cite{faruqui2016problems}, word similarity remains the most frequently used method of evaluation in the literature. There are multiple test sets available but in almost all of them word pairs are considered out of context. 
To the best of our knowledge, the only word similarity data set providing word context is SCWS \cite{huang2012improving}.  
To evaluate our models on SCWS, we run HDP on the data treating each word's context as a separate document.
We compute the similarity of each word pair as follows:
\begin{align*}\small
Sim(w_1, w_2) &= cos(\vt{h}(w_1), \vt{h}(w_2)) \nonumber
\end{align*}
where $\vt{h}(w_i)$ refers to any of the topic-sensitive representations defined in Section~\ref{sect:model}. 
Note that $w_1$ and $w_2$ can refer to the same word.

\begin{table}[t!]
\centering
\small
\begin{tabular}{lccc}
                            & \multicolumn{3}{c}{Dimension} \\ \toprule
      Model        & 100         & 300         & 600         \\ \midrule
SGE + C  \cite{mikolov2013efficient}  &     0.59        &    0.59         &      0.62       \\
MSSG \cite{neelakantan2014efficient} & 0.60 & 0.61  & \textbf{0.64} \\\cmidrule(r){1-4}
HTLE          &     \textbf{0.63}        &       0.56      &       0.55      \\
HTLEadd              &         0.61    &      \textbf{0.61}       &      0.58       \\
STLE                      &     0.59        &      0.58       &       0.55      \\ \bottomrule
\end{tabular}
\caption{Spearman's rank correlation performance for the Word Similarity task on SCWS.}
\label{scws}
\end{table}
 
Table~\ref{scws} provides the Spearman's correlation scores for different models against the human ranking. We see that with dimensions 100 and 300, two of our models obtain improvements over the baseline. The MSSG model of \newcite{neelakantan2014efficient} performs only slightly better than our HLTE model by requiring considerably more parameters (600 vs.\ 100 embedding size). 
  
  \setlength{\tabcolsep}{3pt}

\begin{table}[htb!]
\centering
\small
\tabcolsep=2.5pt%
\begin{tabular}{@{}lclll@{}|lll@{}}
                           && \multicolumn{3}{c}{LS-SE07}  & \multicolumn{3}{c}{LS-CIC} \\ 
                           \toprule
                                                      && \multicolumn{3}{c|}{Dimension}  & \multicolumn{3}{c}{Dimension} \\ 
              Model      & Infer.       & 100         & 300         & 600       & 100         & 300         & 600     \\ \midrule
SGE  &   \multirow{3}{*}{n/a}  & 36.2          &      40.5       &      41.1   &      30.4     &       32.1      &       32.3       \\
SGE + C  &  &      36.6       &       40.9      &     41.6   &  32.8      &    36.1         &      36.8         \\
MSSG & & 37.8 & 41.1 & 42.9 & 33.9 & 37.8 &  39.1 \\
\midrule
HTLE    & \multirow{3}{*}{Smp} & 39.8$^\blacktriangle$   & 42.5$^\blacktriangle$  &   43.0$^\blacktriangle$  & 32.1 & 32.7 &  33.0   \\
HTLEadd  &   & 39.4$^\vartriangle$   & 41.3$^\blacktriangle$  &   41.8  & 30.4 & 31.5 &    31.7   \\
STLE     &   & 35.2  & 36.7 & 39.0    & 32.9 & 32.3 &  33.9  \\ 
\midrule
HTLE      &  \multirow{3}{*}{Exp}   &     \textbf{40.3}$^\blacktriangle$       &      \textbf{42.8}$^\blacktriangle$       &      \textbf{43.4}$^\blacktriangle$     &     36.6$^\blacktriangle$       &      \textbf{40.9}$^\blacktriangle$        &      \textbf{41.3}$^\blacktriangle$      \\
HTLEadd  &                &       39.9$^\blacktriangle$     &       41.8$^\blacktriangle$       &      42.2       &        35.5$^\vartriangle$     &  37.9$^\vartriangle$          &      38.6      \\
STLE           &     &       38.7$^\vartriangle$      &     41.0         &      41.0       &      \textbf{36.8}$^\blacktriangle$      &  36.8   &     37.1      \\ \bottomrule
\end{tabular}
\caption{\label{embs_tables00}GAP scores on LS-SE07 and LS-CIC sets. For SGE + C we use the \textit{context} embeddings to disambiguate the substitutions. Improvements over the best baseline (MSSG) are marked $^\blacktriangle$ at $p<.01$ and $^\vartriangle$ at $p < .05$.} 
\end{table}

\begin{table*}[h!]
\centering
\small
\setlength{\tabcolsep}{.3em}
\begin{tabular}{l|ccccl|ccccl|ccccl}
\toprule
     \multirow{2}{*}{Model}    &  \multicolumn{4}{c|}{Dim = 100}                                                                 &       & \multicolumn{4}{c|}{Dim = 300}                                            &       & \multicolumn{4}{c|}{Dim = 600}                                                                 &       \\
        & n.             & v.             & adj.        & \multicolumn{1}{c|}{adv.}           & All   & n.             & v.  & adj.        & \multicolumn{1}{c|}{adv.} & All   & n.             & v.             & adj.        & \multicolumn{1}{c|}{adv.}           & All   \\ \midrule
SGE + C  & 37.2            & 31.6            & 37.1            & \multicolumn{1}{c|}{42.2}            & 36.6 & 39.2            & 35.0 & 39.0            & \multicolumn{1}{c|}{\textbf{55.4}}  & 40.9 & 39.7            & 35.7            & 39.9            & \multicolumn{1}{c|}{\textbf{56.2}} & 41.6 \\
HTLE     & \textbf{42.4} & \textbf{33.9} & \textbf{38.1} & \multicolumn{1}{c|}{\textbf{49.7}} & \textbf{40.3} & \textbf{44.9} & \textbf{37.0} & \textbf{41.0} & \multicolumn{1}{c|}{50.9}  & \textbf{42.8} & \textbf{45.2} & \textbf{37.2} & \textbf{42.1} & \multicolumn{1}{c|}{51.9}            & \textbf{43.4}\\ \bottomrule
\end{tabular}
\caption{GAP scores on the candidate ranking task on LS-SE07 for different part-of-speech categories.}
\label{nvaa}
\end{table*} 

\subsection{Lexical Substitution Task}

This task requires one to identify the best replacements for a word in a sentential context.  
The presence of many polysemous target words makes this task more suitable for evaluating sense embedding. 
Following \newcite{melamud2015simple} we pool substitutions  
from different instances and rank them by the number of annotators that selected them for a given context. 
We use two evaluation sets: LS-SE07 \cite{mccarthy2007semeval}, and LS-CIC \cite{kremer2014substitutes}.

Unlike previous work \cite{DBLP:conf/emnlp/SzarvasBH13,kremer2014substitutes,melamud2015simple} we do not use any syntactic information, 
motivated by the fact that high-quality parsers are not available for most languages.
The evaluation is performed by computing the Generalized Average Precision (GAP) score \cite{kishida2005property}. 
We run HDP on the evaluation set  
and compute the similarity between target word $w_t$ and each substitution $w_s$ using two different inference methods in line with how we incorporate topics during training:

\begin{equation*}
\small
\begin{split}
\textrm{\textbf{Sampled (Smp):}} &\ Sim_\textrm{TSE}(w_s, w_t) = \\
\ cos(\vt{h}(w_s^\tau), \vt{h}(w_t^{\tau'})) &+  \frac{\sum_{c}  cos(\vt{h}(w_s^{\tau}), \vt{o}(w_c)) }{C} \nonumber\\ 
\textrm{\textbf{Expected (Exp):}} &\ Sim_\textrm{TSE}(w_s, w_t) = \sum_{\tau, \tau'}  \ p(\tau) \ p(\tau') \\ 
\ cos(\vt{h}(w_s^\tau), \vt{h}(w_t^{\tau'})) &+  \frac{\sum_{\tau,c}  cos(\vt{h}(w_s^{\tau}), \vt{o}(w_c)) \ p(\tau) }{C}  \nonumber
\end{split}
\end{equation*}
where $\vt{h}(w_s^\tau)$ and $\vt{h}(w_t^{\tau'})$ are the representations for substitution word $s$ with topic $\tau$ and target word $t$ with topic $\tau'$ respectively (cf. Section~\ref{sect:model}),
$w_c$ are context words of $w_t$ taken from a sliding window of the same size as the embeddings,
$\vt{o}(w_c)$ is the context (i.e., output) representation of $w_c$, and $C$ is the total number of context words.
Note that these two methods are consistent with how we train HTLE and STLE.

The \textit{sampled} method, similar to HTLE, uses the HDP model to assign topics to word occurrences during testing. The \textit{expected} method, similar to STLE, uses the HDP model to learn the probability distribution of topics of the context sentence and uses the entire distribution to compute the similarity. 
For the Skipgram baseline we compute the similarity $Sim_\textrm{SGE+C}(w_s, w_t) $ as follows:
\begin{align*}\small
\; cos(\vt{h}(w_s), \vt{h}(w_t)) +\frac{\sum_{c} cos(\vt{h}(w_s), \vt{o}(w_c))}{C} \nonumber
\end{align*}
which uses the similarity between the substitution word and all words in the context, as well as the similarity of target and substitution words.

Table~\ref{embs_tables00} shows the GAP scores of our models and baselines.\footnote{We use the nonparametric rank-based Mann-Whitney-Wilcoxon test 
\cite{sprent2016applied} to check for statistically significant differences between runs.}  
One can see that all models using multiple embeddings per word perform better than SGE. Our proposed models outperform both SGE and MSSG in both evaluation sets, with more pronounced improvements for LS-CIC. 
We further observe that our \textit{expected} method is more robust and performs better for all embedding sizes.   

Table~\ref{nvaa} shows the GAP scores broken down by the main word classes: noun, verb, adjective, and adverb. With 100 dimensions our best model (HTLE) yields improvements across all POS tags, with the largest improvements for adverbs and smallest improvements for adjectives. When increasing the dimension size of embeddings the improvements hold up for all POS tags apart from adverbs. It can be inferred that larger dimension sizes capture semantic similarities for adverbs and context words better than other parts-of-speech categories. 
Additionally, we observe for both evaluation sets that the improvements are preserved when increasing the embedding size. 

\section{Related Work}

While the most commonly used approaches learn one embedding per word type \cite{mikolov2013efficient,pennington2014glove}, 
recent studies have focused on learning multiple embeddings per word due to the ambiguous nature of language \cite{qiu-tu-yu:2016:EMNLP2016}.
\newcite{huang2012improving} cluster word contexts and use the average embedding of each cluster as word sense embeddings, which yields improvements on a word similarity task. \newcite{neelakantan2014efficient} propose two approaches, both based on clustering word contexts: In the first, they fix the number of senses manually, and in the second, they use an ad-hoc greedy procedure that allocates a new representation to a word if existing representations explain the context below a certain threshold.
\newcite{li-jurafsky:2015:EMNLP} used a CRP model to distinguish between senses of words and train vectors for senses, where the number of senses is not fixed.  
They use two heuristic approaches for assigning senses in a context: `greedy' which assigns the locally optimum sense label to each word, and `expectation' which computes the expected value for a word in a given context with probabilities for each possible sense. 

\section{Conclusions}

We have introduced an approach to learn topic-sensitive word representations that exploits the document-level context of words and does not require annotated data or linguistic resources. 
Our evaluation on the lexical substitution task suggests that  
topic distributions capture word senses to some extent. 
Moreover, we obtain statistically significant improvements in the lexical substitution task while not using any syntactic information.  
The best results are achieved by our hard topic-labeled model which learns topic-sensitive representations by assigning topics to target words. 

\section*{Acknowledgments}

This research was funded in part by the Netherlands Organization for Scientific Research (NWO) under project numbers 639.022.213 and 639.021.646, and a Google Faculty Research Award. We thank the anonymous reviewers for their helpful comments.

\bibliography{acl2017}
\bibliographystyle{acl_natbib}

\end{document}